\documentclass[journal]{IEEEtran}
\ifCLASSINFOpdf
\else
\fi
\usepackage{amsmath,graphicx}
\usepackage[colorlinks=true,linkcolor=blue,citecolor=blue]{hyperref}
\usepackage{amsmath}
\usepackage{amssymb}
\usepackage{url}
\usepackage{colortbl} 
\usepackage{arydshln} 
\usepackage{multirow} 
\usepackage{multicol}
\usepackage{verbatim}

\begin{document}

\title{Crowd Scene Analysis by Output Encoding}
%\title{Accurate Crowd Localization by Output Encoding}


\author{Yao~Xue,~
		Siming~Liu,~
		Yonghui~Li,~
        Xueming~Qian,~\IEEEmembership{Member,~IEEE,}
        % <-this % stops a space
\thanks{
	
%Y. Xue (yxue2@ualberta.ca) and X. Qian (qianxm@mail.xjtu.edu.cn) are with the Key Laboratory for Intelligent Networks and Network Security, Ministry of Education, and the Smiles Laboratory, Xi'an Jiaotong University, Xi'an 710049, China.

Yao Xue (yxue2@ualberta.ca), Siming Liu, Yonghui Li are with the Faculty of Electronics and Information Engineering, Xian Jiaotong University, Xi’an, Shaanxi 710049, China. Xueming Qian (corresponding author, qianxm@mail.xjtu.edu.cn) is with the Ministry of Education Key Laboratory for Intelligent Networks and Network Security, School of Information and Communication Engineering, and  SMILES LAB, Xi’an Jiaotong University, Xi’an 710049, China. This work was supported in part by the NSFC under Grant 61772407 and 61732008.

}% <-this % stops a space
%\thanks{Manuscript received April 19, 2019; revised August 26, 2019.}
}

\maketitle

\begin{abstract}
Crowd scene analysis receives growing attention due to its wide applications. Grasping the accurate crowd location (rather than merely crowd count) is important for spatially identifying high-risk regions in congested scenes. In this paper, we propose a Compressed Sensing based Output Encoding (CSOE) scheme, which casts detecting pixel coordinates of small objects into a task of signal regression in encoding signal space. CSOE helps to boost localization performance in circumstances where targets are highly crowded without huge scale variation. In addition, proper receptive field sizes are crucial for crowd analysis due to human size variations. We create Multiple Dilated Convolution Branches (MDCB) that offers a set of different receptive field sizes, to improve localization accuracy when objects sizes change drastically in an image. Also, we develop an Adaptive Receptive Field Weighting (ARFW) module, which further deals with scale variation issue by adaptively emphasizing informative channels that have proper receptive field size. Experiments demonstrate the effectiveness of the proposed method, which achieves state-of-the-art performance across four mainstream datasets, especially achieves excellent results in highly crowded scenes. More importantly, experiments support our insights that it is crucial to tackle target size variation issue in crowd analysis task, and casting crowd localization as regression in encoding signal space is quite effective for crowd analysis.
\end{abstract}


% \fi
\IEEEpeerreviewmaketitle

\section{Introduction}

\IEEEPARstart{T}{he} wide deployment of surveillance cameras in many cities stimulates the recent research interests in visual analysis of crowd scenes. It has a wide range of real-world applications, such as crowd surveillance, traffic monitoring and planning, even cell counting. %Major challenges for accurate crowd counting consist of object occlusion, complex background, annotation errors, density variations.

Mainstream approaches can be summarized into two categories: counting by density prediction and counting by detection. State-of-the-art methods use regression-based models (\emph{e.g.} object density estimator), which explicitly learn to count the objects of interest. These counting by density prediction approaches \cite{Switch-CNN} \cite{RAZ} \cite{ShanghaiTech} \cite{spatial} \cite{SANet} \cite{Lu2018Crowd} have achieved superior performance on several existing counting datasets \cite{ShanghaiTech} \cite{WorldExpo} \cite{UCF-QNRF}.

%\textbf{Density Deviation of Regression-based Approaches}

Density prediction methods measure deviation of the output density from a ground truth density during their training process. In order to train the density predictor, one has to create ground truth density map by smoothing on point (people head) annotations. This smoothing operation is extremely sensitive to high crowd density. If objects are densely present, peaks in the density map tend to merge. Neighboring peaks in the density map are very easy to mix together, thereby introducing errors in the very beginning stage. Additionally, sparse object locations create an imbalance in the cost function between positive and negative samples.

\begin{figure}[h]
	\centering
	\includegraphics[width=9cm]{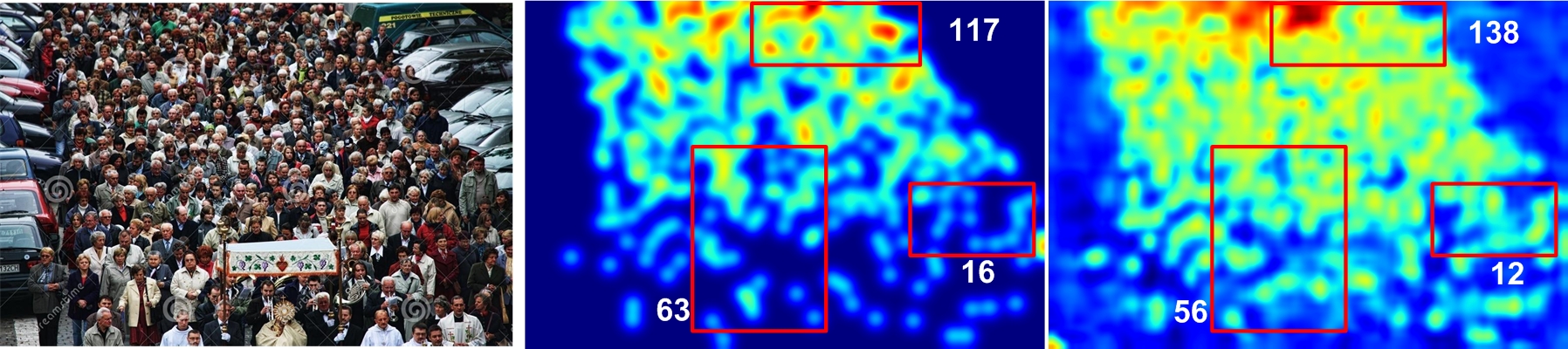}
	\caption{Examples of predicted density maps for the ShanghaiTech dataset (true count: 361 prediction: 365). Left column: crowd image. Middle column: ground truth. Right column: prediction.}
	\label{intro-1}
\end{figure}

Recent research \cite{RAZ} has indicated that only predicting object count or global density map for congested scenes is insufficient for real-world demand such as public safety or traffic flow monitoring. Grasping the accurate crowd position (rather than merely global density) is important for spatially identifying high-risk regions from whole monitor images. But there was a recent fashion to perform counting by density prediction. Although the crowd count could be a precise estimation in the whole image level, the predicted density map can largely deviate from the true density map in sub-image level. We illustrate this phenomenon in Fig.~\ref{intro-1}, where the global true count: 361 and estimated count: 365 are quite close to each other. But the estimated density does not offer reliable approximations to ground truth in specific image regions, whose true counts are 117, 63 and 16, but estimated counts are 138, 56 and 12.

%Reasons behind density deviation can be summarized into the following aspects.
%目前这个行文写的挺好的，别往那三点结构靠！

%\textbf{Limitation of Localization in Pixel Space}

Despite the fashion of counting by density prediction, many researches \cite{RAZ} \cite{UCF-QNRF} \cite{Blobs-Localization} \cite{BoxOut} still propose to fulfill crowd counting and crowd localization simultaneously. Recent researches \cite{UCF-QNRF} \cite{RAZ} have argued that tackling localization task can bring noteworthy benefits for counting task, such as counting error correction, enabling localization-based applications \emph{e.g.} human tracking. Counting by localization approaches formulate the counting task as a classic computer vision problem: object detection. For crowd scenes with few occlusion and low density, well-trained detector is able to localize objects, then object count is naturally obtained. These methods strive to directly predict the pixel level {x,y}-coordinate of objects. In this way, inevitable system prediction errors will also directly affect the pixel level coordinates of detected objects, such as position shift or size scaling of bounding boxes or dotted annotations. For general object detection, position shift or size scaling \emph{e.g.} in tens pixels could be acceptable. However for accurate crowd localization, such shift or scaling can pose much bigger challenges for localization in crowded scenes where objects are densely distributed, and would result in complete false detections. It is often seen that the more crowded an area is, the more inaccurate detection could emerge.

%\textbf{Compressed Sensing based Output Encoding}

In this paper, we treat the task of crowd localization as an application of integrating Compressed Sensing based Output Encoding (CSOE) with supervised learning by CNN. As the output space is sparse for the crowd localization problem (only a few pixel locations are people head centroids), we can employ CSOE here. Furthermore, CS theory dictates that pairwise distances in the sparse space are approximately maintained in the compressed space \cite{CS}. So, even after the output space encoding, CNN still targets the original output space in an equivalent distance norm.

The principle behind our CSOE module is straightforward. CS converts the sparse output pixel space into dense and short vectors. As a regressor, we use a trained CNN to predict the compressed vectors. Then using a reconstruction algorithm, we recover sparse cell locations in the output pixel space. In other words, we seek a different route that casts the problem of detecting variable number of small objects into a task of signal regression in encoding signal space. Compared to pixel coordinates representation of crowd in images, a signal representation is more robust to inevitable system errors.

On the basis of CSOE, we further render the structure of CNN+CS end-to-end trainable. CS-based encoding started with the work of Hsu et al. \cite{CS} that proved a generalization prediction error bound. The error bound depends on two factors. How well the machine learner has predicted; and how well the recovery process has worked. In this work, we realize the joint optimization of both the machine learner and the recovery process by implementing them as CNN-based observation layers and sparse coding based reconstruction layers of the end-to-end trainable network. In addition, we derive a backpropagation rule for the reconstruction layers. Thus, the end-to-end training process is not only occurring within the observation layers, but also back-propagates error signals to optimize the parameters of the reconstruction layers, finally removes the risk of gradient vanishing in the deep reconstruction layers. This is different from the conventional sequential pipeline, where each component is optimized independently and could cause error accumulation.

Scale variation is a crucial challenge for object localization. An solution is to deploy in-network feature pyramids. E.g. FPN \cite{FPN} adds a top-down connection to incorporate semantic high level features. Facing the issue of scale variation, we create Multiple Dilated Conv Branches (MDCB) sharing convolution weight but having different receptive field sizes for objects in different sizes. Furthermore, we deploy center pooling \cite{CenterNet} to introduce the visual patterns within objects into the centroid point detection process.

Due to the distance, capture angle between camera and crowd, targets present huge size variation. Recent crowd analysis works \cite{ShanghaiTech} \cite{CSRNet} \cite{SEnet} \cite{TridentNet} have suggested that the receptive field sizes of neural network should not be fixed, but modulated by the stimulus. Unfortunately, this property does not receive much attention in constructing deep learning models. In the paper, we present a nonlinear approach to aggregate information from multiple kernels to realize the adaptive changing of receptive field sizes. Specifically, we introduce an Adaptive Receptive Field Weighting (ARFW) module, which consists of a triplet of operations: spatially Aggregate, inter-channel Weight and Modulate. On the basis of backbone that generates multiple branches with various kernel sizes corresponding to different receptive field sizes, the Aggregate operator produces a channel descriptor by aggregating feature maps across their spatial dimensions. This descriptor enables a global receptive field within channel-wise feature maps. The Weight operator is based on two fully connected layers and produces a set of weights between channels. The Modulate operator modulates the feature maps of different receptive field sizes according to the channel weights. To this end, we propose a mechanism through which networks can learn to use global information to adaptively emphasize informative channels that have proper receptive field size and suppress less useful ones.

%\begin{figure}[h]
%	\centering
%	\includegraphics[width=9cm]{intro-2.jpg}
%	\caption{purple still have 10+ pixel error.}
%	\label{intro-2}
%\end{figure}

%To illustrate this phenomenon, we show an example in Fig.~\ref{intro-2}. A coordinates error is defined as the  equaling to localization results on the image. The red / green dots denote true / predicted human heads respectively. Counting by detection approaches suffer more from a loss of detection accuracy especially in the upper area of the image, where people are more densely present.

%Another advantage is that end-to-end training avoids non-trivial pre-processing or post-processing operations, which requires significant prior knowledge about the task and the dataset.

The \emph{contributions} of this paper are summarized as. (1) We propose the Compressed Sensing based Output Encoding (CSOE) scheme, which casts object localization as signal regression task, CSOE helps to boost localization performance in circumstances where targets are highly crowded without huge scale variation. (2) We create Multiple Dilated Convolution Branches (MDCB) that aims to improve localization accuracy when objects sizes change drastically in an image and offers a set of different receptive field sizes. Unlike traditional convolution+pooling operations, MDCB avoids excessive loss of detail resolution which poses huge challenges to high density crowd scenes analysis. (3) To further deals with scale variation issue, we propose an Adaptive Receptive Field Weighting (ARFW) mechanism through which networks can learn to use global information to adaptively emphasize informative channels that have proper receptive field size. (4) In the observation head, we enrich geometric center information by center pooling to capture more recognizable visual patterns that locate within objects, while may not always lie on the geometric center of objects. (5) We render the method end-to-end trainable by deriving an independent backpropagation rule for the reconstruction layers to prevent gradient vanishing and error accumulation brought by conventional cascaded networks.

\begin{figure*}[htbp]
	\centering
	\includegraphics[width=17cm]{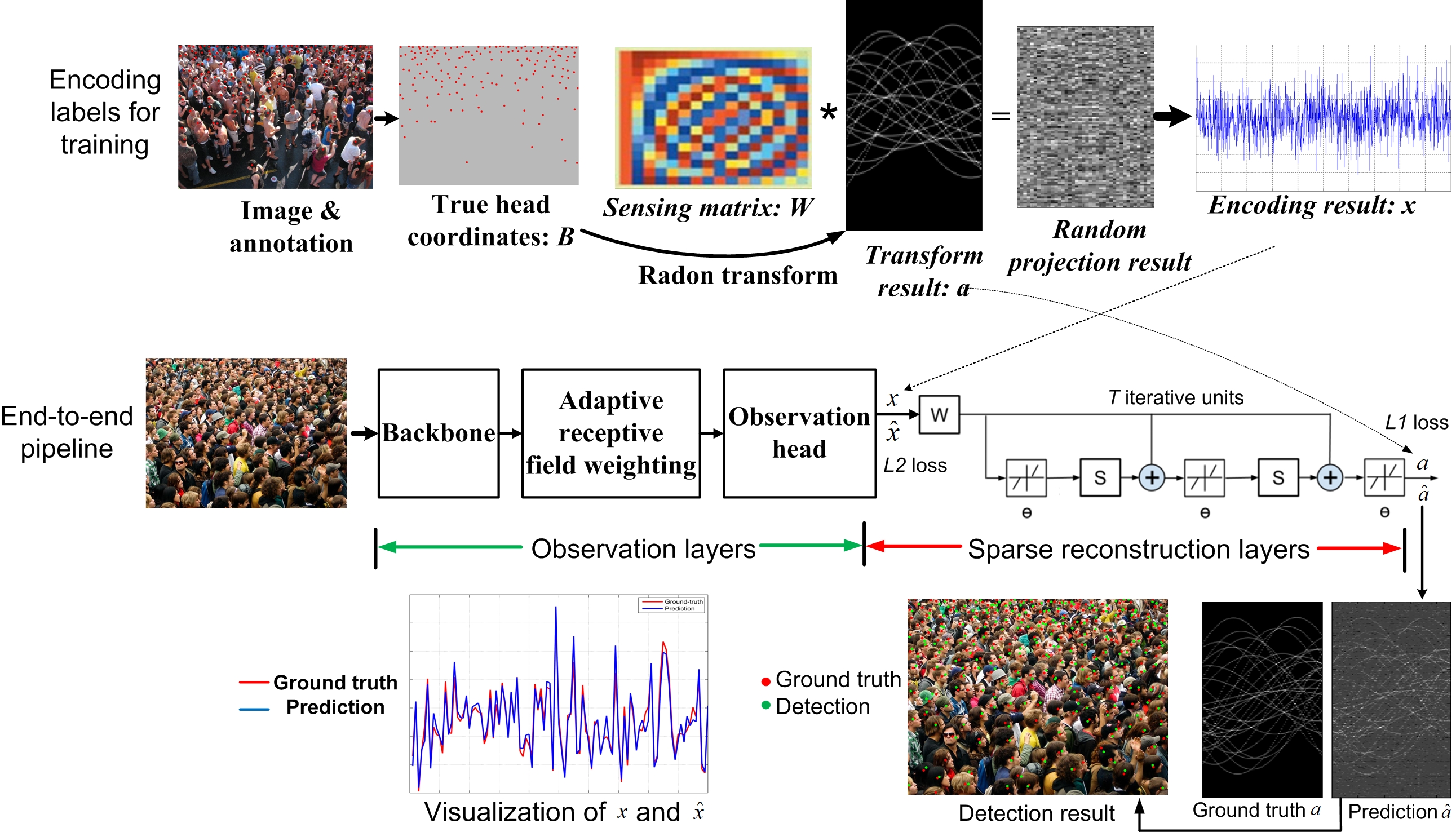}
	\caption{System overview of the proposed method.}
	\label{system}
\end{figure*}

\section{Related work}

\emph{Density prediction approaches}
%如果觉得此节单薄，可从Archor文的2.2节翻译继续加

Regression-based algorithms are developed to regress the object count of interest from crowd images. Recently, end-to-end trainable deep neural networks (DNNs) are widely adopted in crowd counting task. These DNNs are optimized to predict a density map that approximates ground truth crowd distribution. For example, Zhang et al. \cite{cross-scene} solve the cross-scene crowd counting problem with a deep convolutional neural network fed with density map and global count datasets. The quality of density map can have a non-negligible impact on counting results. \cite{Adaptive} witnesses a significant  boost in the crowd counting performance after adopting density map refinement framework.

\emph{Localization approaches}

Early counting by detection methods \cite{Early1} \cite{Early2} \cite{Early3} \cite{Early4} rely on hand-crafted features, which cannot well handle those highly congested scenes with occlusions. Nowadays deep learning models have become key solution to object detection problem. Therefore, many researches seek to apply deep learning based detection frameworks into crowd counting, and have achieved remarkable performance improvements. \cite{Archor28} proposes an end-to-end trainable human detector for crowded scenes. For scenes with few occlusion and low density, well-trained detector is able to localize objects. For example, DecideNet \cite{DecideNet} deploys human detector to rectify counting in low density regions. However in many crowd scenes, people head are so small that bounding box annotations are not suitable. Therefore, dot annotation on every head center is usually used in crowd localization task. This limits the application of detection based methods in crowd counting. Additionally, the high density becomes a hurdle for traditional counting by detection approaches. In this paper, we will demonstrate that detection based method can not only give highly precise localization results, but also obtain comparable even better counting performance by introducing compressed sensing techniques into detection framework.

\emph{Compressed sensing based output encoding}

Compressed sensing (CS) \cite{Alpher17} \cite{Alpher18} \cite{Alpher16} and sparse coding (SC) \cite{Elad2010} have emerged as new frameworks for signal acquisition and reconstruction, with rich theoretical results and significant practical applications, such as MRI scan time reduction \cite{BirnsKKSN16} and economical camera design \cite{Duarte2008}. CS-based encoding has a rather modest presence in the literature, where it was applied with linear and non-linear machine learners. One early research in the output encoding with error correcting ability \cite{ECOC} had shown superior accuracy. In the recent past, redundancy in the output representation \cite{RAkEL} yielded more accurate predictions. Recently, non-linear predictors such as a Bayesian learner \cite{Bayesian-CS}, Decision trees \cite{output-space-thesis} or CNN \cite{my-TMI} were used. Viswanathan et al. \cite{Bayesian-CS} used Bayesian inference with CS and showed good accuracy in prediction. Decision trees and gradient boosting had also been used in conjunction with CS encoding to yield good prediction accuracies \cite{output-space-thesis}. Recent researches \cite{my-TMI} \cite{ADMM-CSNet} \cite{ConvCSNet} focused on the cross domain application of compressive sensing and deep learning. For example, \cite{my-TMI} develops a CS-based tumor cell localization scheme and proposing an end-to-end training network. However unlike crowd analysis, size variations of tumor cells are much smaller due to strict operating conditions during making medical microscopy slides, such as tissue collection, sectioning, staining, scanning, etc. Consequently, \cite{my-TMI} doesn't pay attention to solve target size variation issue. Without the ability to offer different receptive field sizes, \cite{my-TMI} also cannot adaptively emphasize informative channels that have proper receptive field size.

\emph{Channel-wise attention mechanisms}

Proper regulation of informative channels can have a positive effect on the overall performance of DNNs. Hu et al. \cite{SEnet} use a lightweight gating mechanism in SENet to adaptively recalibrate feature maps based on channel-wise dependencies. SKNet \cite{SKNet} exploits the aggregation of the feature maps of different-sized kernels via selection weights to self-modulate receptive Field sizes and achieves superior performance in object recognition. \cite{Dropout} proposes weighted channel dropout to filter channels in accordance to activation status and elevates detection performance with a slight computational cost.

\emph{Multi-scale architectures}

A number of research attentions have been paid to the scale variation issue. Several piece of initial works \cite{image-pyramid-1} \cite{image-pyramid-2} \cite{image-pyramid-3} deploy multi-scale image pyramid to refine counting performance in areas that objects are densely present. Recent progress have taken spatial locality \cite{spatial}, cross scale aggregation \cite{SANet}, and adaptive scale \cite{Lu2018Crowd} into consideration. It has been proven effective to gather multiple branches with distinct targets, such as Switch-CNN \cite{Switch-CNN} and RAZ-Net \cite{RAZ}. \cite{ShanghaiTech}  develops a multi-column CNN that uses different convolution kernel sizes to deal with varying density. \cite{top-down} adopts a scale-aware training scheme for the multi-branch architecture to give each branch a specialty for corresponding scales and achieves remarkable improvements over baseline approach. Additionally, enlarging receptive field of deep network is another insightful idea, for example CSRnet \cite{CSRNet} deploys a sequence of dilation convolutions and takes human body structure into consideration.

\section{The proposed method}

\subsection{System Overview}

The proposed detection framework consists of two components: (1) a crowd location encoding scheme based on compressed sensing, (2) an end-to-end trainable network which is made up of observation layers and sparse reconstruction layers. The structure of the whole framework is shown in Fig.~\ref{system}.

To encode training labels, We propose a crowd location encoding scheme, which converts people location from pixel space representation to compressed signal representation. Then, each training pair, consisting of a crowd image and the signal, trains a CNN to work as a multi-label regression model. We employ a joint loss function during training, because it is suitable for both signal regression and signal reconstruction. During testing, the observation layers of the network predicts crowd location signal for each test image. After that, sparse reconstruction layers of the network predicts the pixel level crowd locations.

\subsection{Crowd Location Encoding Scheme}

Our proposed method relies on encoding people head locations into a dense code that a CNN predicts from an input image. We use a form of encoding that we refer to as encoding by Radon transform \cite{radon} followed by a random projection \cite{AAAI_workshop}. Radon transform is often seen as a mapping from Cartesian rectangular coordinates to polar coordinates, and is widely used for image reconstruction from the projections associated with cross-sectional scans of an object \cite{radon}.

Referring to the encoding method shown in Fig.~\ref{system}, $B$ denotes the binary (0/1) ground truth head location matrix with size $h \times w$. In the first step of the encoding method, $B$ is converted to another sparse matrix $a$ by Radon transform. Radon transform projects $B$ along a radial line oriented at a specific angle. Here we use $r$ angles uniformly varying in the range of [0, 179] degrees. The transform results in matrix $a$ with size $n \times r$, with $n=\sqrt[]{h^2+w^2}$.

Since radon transform \cite{radon} of people head locations is a sparse signal. In the second step of the encoding method, we apply CS-based encoding of the people head locations that compresses a sparse vector $a$ into a much smaller denser vector $x$ with a sensing matrix $D$ by:
\begin{equation} \label{eq:cs}
x= D a,
\end{equation}
where $D$ is a $m \times n$ random Gaussian sensing matrix (each element is independently and identically distributed zero mean Gaussian with variance $1/m$), with typically $m \ll n.$ CS theory \cite{Alpher17,Alpher18} states that given $x$ and $D$, a convex optimization can recover $a,$ provided the sensing matrix $D$ satisfies a restricted isometry property (RIP) and $m \geq  C_m k log(n)$, where $C_m$ is a small constant greater than one and $k$ is the maximum number of non-zero elements in $a.$ 

Given $D$ and $x$, the recovery of $a$ typically relies on a convex optimization with a penalty expressed by $L_1$ norm:
\begin{equation} \label{eq:sp1}
\underset{a}{\text{min}} \ \frac{1}{2}\| Da-x \|_2^{2}+\lambda\| a \|_1, %\ \ \ \text{w.r.t} \ \ u=Dv
\end{equation}
where $\lambda$ is a non-negative weight balancing the two terms in the cost function (\ref{eq:sp1}). Various algorithms exist today that can optimize (\ref{eq:sp1}). Examples include orthogonal matching pursuit (OMP) \cite{OMP} and dual augmented Lagrangian (DAL) \cite{DAL}. In this work, we realize the recovery process by an end-to-end neural network structure. Details can be found in section \emph{D}.

\subsection{Signal Regression by Observation Layers}

We utilize CNN to build a regression model between a crowd image and its people head location signal $x$.

\subsubsection{Backbone}

We adopt the truncated VGG-16 network, i.e. the first 13 layers of VGG-16, as the input structure of our backbone, such truncated network has shown superior transfer ability for crowd analysis (\cite{RAZ}, \cite{CSRNet}). VGG-16 model pre-trained on ImageNet is used to initialize the backbone.

\begin{figure}[htbp]
	\centering
	\includegraphics[width=8cm]{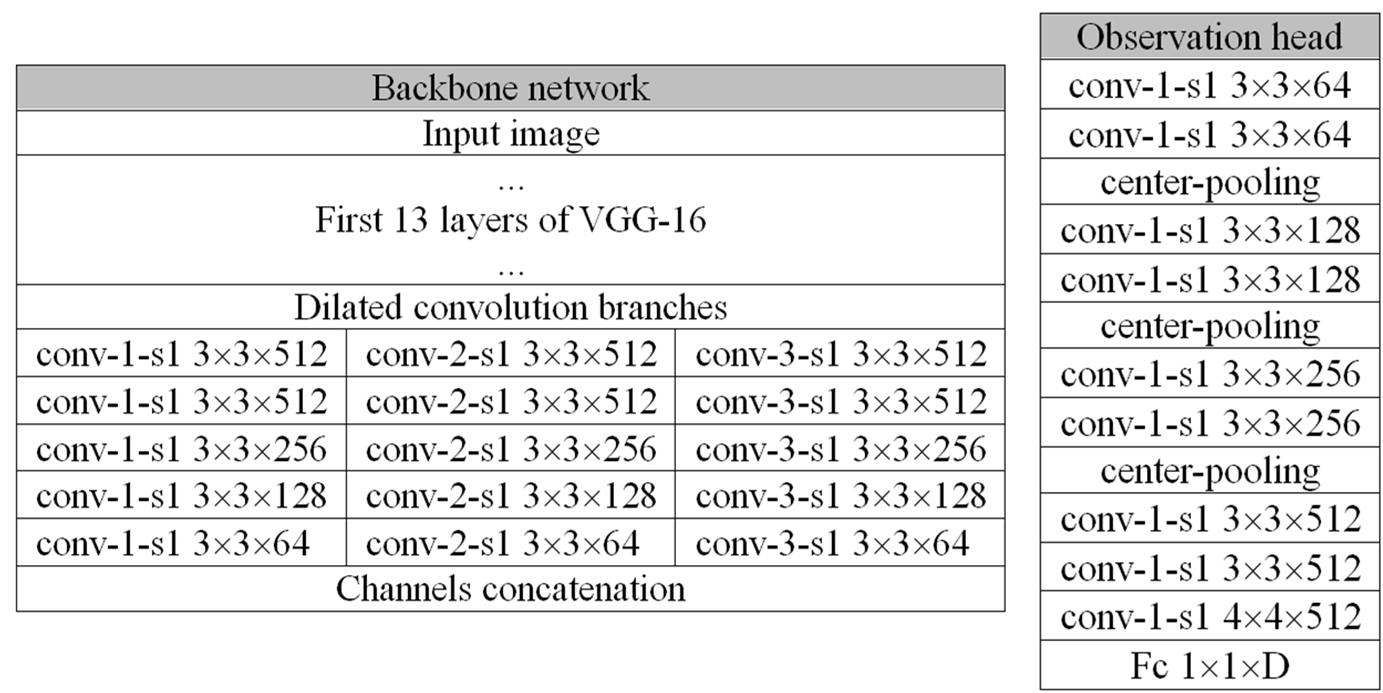}
	\caption{Configuration of Backbone and Observation head. All convolutional layers use padding. Convolution layer parameters are denoted as "Conv-(dilation rate)-stride kernel $\times$ kernel $\times$ filters". Center-pooling layer are conducted over a 3 $\times$ 3 pixel window with stride 2.}
	\label{backbone-head}
\end{figure}

To equip the backbone with different receptive field sizes, following the truncated VGG-16, we create Multiple Dilated Convolution Branches (MDCB) architecture with different dilation rates to adapt the receptive fields for objects of different scales. To control the receptive field of the backbone,  we use different dilation rates which vary from 1 to 3 for a 3 $\times$ 3 convolutions. As the output of the backbone, we perform channels concatenation to splice the feature maps from different channels to form a set of feature responses with different receptive field sizes. An illustration of our backbone is shown in the left part of Fig.~\ref{backbone-head}.

\begin{figure}[htbp]
	\centering
	\includegraphics[width=10cm]{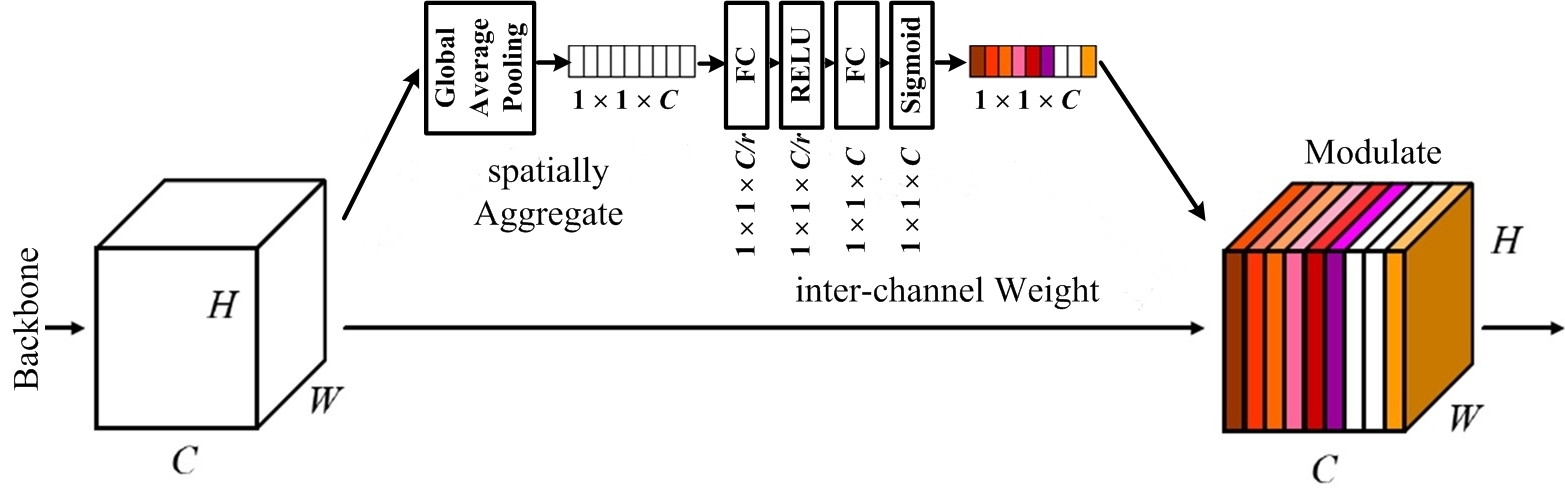}
	\caption{Illustration of Adaptive Receptive Field Weighting (ARFW) block.}
	\label{weight-channel}
\end{figure}

A risk of our multi-branch block is that it introduces many parameters which may potentially cause overfitting. To prevent this risk, we make different branches share the same structure and weights, and only vary the dilation rate between branches. The advantages of weight sharing are three aspects. It doesn't need extra parameters compared with the original backbone network. Second, it reflects our motivation that objects in different sizes should be processed by an uniform transformation only with different scales. Finally, in this way the same set of parameters are fully trained for different scale ranges under different receptive fields.

\subsubsection{Adaptive Receptive Field Weighting}

Adaptive Receptive Field Weighting contains the following three modules.

\textbf{Aggregate: Spatial Information Embedding}

In order to exploit channel dependencies, we first consider the signal to each channel in the output features. Each learned filters operates with a local receptive field and consequently can only exploit contextual information within its receptive field. To tackle this issue, we propose to spatially aggregate feature maps across their spatial dimensions, so that global information is embedded into a channel descriptor. To do that, we use global average pooling to generate channel-wise statistics. A statistic $a$ is generated by shrinking V through its spatial dimensions $H \times W$, so that the $c$-th element of $a$ is computed by:
\begin{equation}
a_{c}=F_{ag}(v_{c})=\dfrac{1}{H \times W}\sum_{i=1}^{H}\sum_{j=1}^{W}v_{c}(i,j)
\end{equation}

\textbf{Adaptive Channel-wise Weighting}

To make use of the information embedded in the aggregate operation, the second operation aims to fully capture channel-wise dependencies. This operation should be capable of learning a nonlinear interaction between channels and allow multiple channels to be emphasized simultaneously rather than enforcing a one-hot activation. To fulfill this objective, we choose to employ a gating mechanism with a sigmoid activation:
\begin{equation}
z=F_{we}(a,W)=\sigma(g(a,W))=\sigma(W_{2}\sigma(W_{1}a))
\end{equation}
where $\sigma$ refers to the ReLU function, $W_{1}$ and $W_{2}$ denote the parameters of the two FC layers. To aid model generalization, we parameterize the gating mechanism by forming a bottleneck with two fully-connected (FC) layers, one is a nonlinear dimensionality-reduction layer with reduction ratio $r$, the other is a dimensionality-increasing layer returning to the number of input channels. Deploying two FCs is beneficial for modeling the complex dependencies between channels, and also helpful for limiting model complexity. This operator maps the input-specific descriptor $a$ to a set of channel weights $z$.

\textbf{Modulate Channels}

The last operator modulates the feature maps of different receptive field sizes according to the channel weights. The final output of the ARFW block is obtained by rescaling $V$ with the activations $z$:
\begin{equation}
\widetilde{u_{c}}=F_{mod}(v_{c},z_{c})=z_{c}v_{c}
\end{equation}
where $F_{mod}(v_{c},z_{c})$ refers to channel-wise multiplication between the scalar $z_{c}$ and the feature map $v_{c}$. ARFW block intrinsically introduce dynamics conditioned on the input, which can be regarded as a self attention function on channels.

\subsubsection{Observation Head}
Observation head takes the weighted feature maps from ARFW as input and then predicts the encoding signal. Besides convolution and fully-connected layers, we further introduce the visual patterns within objects into the center point detection process by using center pooling \cite{CenterNet}. Given a feature map as input, to determine if a pixel in the feature map is a center point, center pooling finds the maximum value in its both horizontal and vertical directions and add them together. By doing this, center pooling helps to enrich center information within objects, since geometric centers of objects cannot always convey recognizable visual patterns. Configuration of the observation head is shown in the right part of Fig.~\ref{backbone-head}.

\subsection{Crowd Localization by Sparse Reconstruction Layers}

Here we present a novel end-to-end trainable network for crowd localization with CS-based output encoding. Given the generalization bound \cite{CS}, optimization of both the prediction and recovery simultaneously in an end-to-end fashion should prove superior.

The bottom diagram of Fig.~\ref{system} shows the end-to-end structure of the network. The input image goes through observation layers composed of a CNN that outputs a dense vector $\hat{x},$ which is compared to the ground-truth dense vector $x$. The predicted dense vector $\hat{x}$ is fed to reconstruction layers that reconstructs a sparse vector $\hat{a},$ which is compared to the ground-truth sparse vector $a$ by $L_1$ norm. Thus, the cost function is a mixture of $L_2$ and $L_1$ norms:
\begin{equation} \label{eq:overall_loss}
loss = \frac{1}{2}\| \hat{x}-x \|_2^{2}+\alpha\| \hat{a}-a \|_1,
\end{equation}
where $\alpha$ is a hyper parameter that balances dense vector errors and sparse vector errors. To optimize the weights in the observation layers and the reconstruction layers jointly, we train the whole model according to the overall loss (\ref{eq:overall_loss}) using gradient descent during backpropagation. 

Suppose $\delta \hat{a}$ and $\delta \hat{x}$ denote the partial derivatives of $L_1$ norm in the loss function (\ref{eq:overall_loss}) with respect to $\hat{a}$ and $\hat{x}$, respectively. Then the following backpropagation rule relates $\delta \hat{a}$ and $\delta \hat{x}$. (Due to space limit, derivation is provided in our additional materials.)

\begin{equation} \label{eq:delx_body}
\delta \hat{x} =D \left(:, p \right)\left[D^{T}D\left( p,p\right)  \right] ^{-1}\delta \hat{a}\left(p \right),
\end{equation}
\begin{equation} \label{eq:delD_body}
\begin{split}
\delta D =\left(x-Da \right) \delta a\left(p \right) ^{T}\left[D^{T}D\left(p,p \right)  \right]^{-1}\\
- D(:,p)\left[D^{T}D\left(p,p \right)  \right]^{-1} \delta a\left(p \right)a\left(p \right)^{T}
\end{split}
\end{equation}

%where $p = \left\{ i:\hat{a}_i\neq0 \right\}$ is the set of indices indicating non-zero components of $\hat{a}.$ $ D(:,p) $ indicates the columns of matrix $D$, whose indices belong to $p$. $D^{T}D\left( p,p\right)$ indicates the principal sub matrix of $D^{T}D$ with column and row indices belonging to $p$. $\delta \hat{a}(p)$ denotes a vector comprising of only those elements of $\delta \hat{a}$ whose indices belong to the set $p.$

The aforementioned rules (\ref{eq:delx_body}), (\ref{eq:delD_body}) may not be numerically stable or efficient for batch training mode, as they involve different matrices to be inverted for different images. We derived an  approximate, numerically stable, and efficient backpropagation for batch training (see additional materials):
\begin{equation} \label{eq:delx_body2}
\delta \hat{x} \approx D \left(:, p \right) \delta \hat{a}\left(p \right),
\end{equation}
\begin{equation} \label{eq:delD_body2}
\delta D \left(:, p \right)\approx \left(\hat{x}-D\hat{a} \right)\delta \hat{a}\left(p \right)^{T} - D \left(:, p \right)\delta \hat{a}\left(p \right) \hat{a}\left(p \right)^{T}. 
\end{equation}

Notice that using a standard toolbox, \emph{e.g.} TensorFlow \cite{Tensorflow} would require both the observation and the reconstruction layers to be differentiable. This requirement brings us to the architecture shown in the bottom diagram of Fig.~\ref{system}. The observation layers being a CNN are differentiable. For the reconstruction layer, we use the differentiable learned iterative shrinkage and thresholding algorithm \cite{LISTA} architecture to compute approximate sparse vectors using a recurrent neural network with a limited number of iterations ($T$). The sparse reconstruction layers have trainable parameters $W=D^T$ and $S=D^TD$. Thus, the entire architecture is now differentiable and end-to-end trainable. We implemented this end-to-end trainable model using TensorFlow \cite{Tensorflow}.

%The matrix $\hat{A}$ is fed to the inverse Radon transform, which outputs a sparse cell location matrix $\hat{B}$. The inverse Radon transform is outside of the end-to-end training in our current work. We plan to include it in the end-to-end training in a future endeavor.

\section{Experiment}

\subsection{Dataset $\&$ Evaluation Metric}

We evaluate the proposed method on four public crowd analysis benchmarks, whose basic information is summarized in Table.~\ref{data-table}. It is necessary to mention that the UCF-QNRF dataset is known as the newest and largest crowd counting and localization dataset, with 1535 high resolution images and over 1.25M annotated heads.

\begin{table}[h] \footnotesize
	\setlength{\abovecaptionskip}{0pt}
	\setlength{\belowcaptionskip}{0pt}
	\begin{center}
		\caption{Summary of evaluation benchmarks.}
		\label{data-table}
		\begin{tabular}{*{22}{c}}
			\hline\noalign{\smallskip}
			Dataset & Resolution & No. of images & Av. Count\\
			\noalign{\smallskip}
			\hline
			\noalign{\smallskip}
			
			ShanghaiTech-A \cite{ShanghaiTech} & 589$\times$868 & 482 & 501\\
			
			ShanghaiTech-B \cite{ShanghaiTech} & 768$\times$1024 & 716 & 123\\
			
			WorldExpo \cite{WorldExpo} & 576$\times$720 & 3980 & 56\\
			
			UCF-QNRF \cite{UCF-QNRF} & 2013$\times$2902 & 1535 & 815\\
			
			\hline
		\end{tabular}
	\end{center}
\end{table}

For crowd counting task, we use the mainstream evaluation metrics: Mean Absolute Error (MAE) and Root Mean Squared Error (RMSE). When the predicted count for image i is $P_{i}$ and the true count $T_{i}$, the MAE and RMSE can be expressed as

\begin{equation}
MAE=\dfrac{1}{n}\sum_{i=1}^{n}|t_{i}-p_{i}|
\end{equation}

\begin{equation}
RMSE=\sqrt{\dfrac{1}{n}\sum_{i=1}^{n}(t_{i}-p_{i})^2}
\end{equation}

For accurate crowd localization task, we adopt the evaluation metric Precision, Recall and $F_1$-score: $F_1=2*Precision*Recall/(Precision+Recall)$ as in \cite{UCF-QNRF}.

\begin{table*}[htbp] \footnotesize
	\setlength{\abovecaptionskip}{0pt}
	\setlength{\belowcaptionskip}{0pt}
	\begin{center}
		\caption{Counting performances on four crowd benchmarks.}
		\label{count-table}
		\begin{tabular}{*{22}{c}}
			
			\hline\noalign{\smallskip}
			
			\multirow{2}{*}{Method} &\multicolumn{2}{c}{ShanghaiTech-A} &\multicolumn{2}{c}{ShanghaiTech-B} &\multicolumn{6}{c}{WorldExpo} &\multicolumn{2}{c}{UCF-QNRF}\\
			
			&MAE &RMSE &MAE &RMSE &S1 &S2 &S3 &S4 &S5 &Ave. &MAE &RMSE\\
			\hline
			\noalign{\smallskip}
			
			CSR-Net \cite{CSRNet} &58.2 &115.0 &10.6 &16.0 &2.9 &11.5 &8.6 &16.6 &3.4 &8.6 &--- &---\\
			
			Cascaded-CNN \cite{Cascaded-CNN} &101.3 &152.4 &20.0 &31.1 &4.8 &32.5 &10.8 &13.3 &4.5 &13.2 &252 &514\\
			
			CP-CNN \cite{CP-CNN} &73.6 &106.4 &20.1 &30.1 &2.9 &14.7 &10.5 &\textbf{10.4} &5.8 &8.9 &--- &---\\
			
			Switch-CNN \cite{Switch-CNN} &90.4 &135.0 &21.6 &33.4 &4.4 &15.7 &10.0 &11.0 &5.9 &9.4 &228 &445\\
			
			MCNN \cite{ShanghaiTech} &110.2 &173.2 &26.4 &41.3 &3.4 &20.6 &12.9 &13.0 &8.1 &11.6 &277 &426\\
			
			SA-Net (patch) \cite{SANet} &67.0 &104.5 &8.4 &\textbf{13.6} &2.6 &13.2 &9.0 &13.3 &3.0 &8.2 &--- &---\\
			
			RAZ-Net \cite{RAZ} &65.1 &106.7 &8.4 &14.1 &\textbf{2.0} &11.8 &9.0 &13.6 &3.3 &8.0 &116 &195\\
			
			ACSCP \cite{ACSCP} &75.7 &102.7 &17.2 &27.4 &2.8 &14.05 &9.6 &8.1 &\textbf{2.9} &	7.5 &--- &---\\
			
			DecideNet \cite{DecideNet} &--- &--- &20.75 &29.42 &\textbf{2.0} &13.14 &8.9 &17.40 &4.75 &9.23 &--- &---\\
			
			Crowd-CNN \cite{Crowd-CNN} &--- &--- &32.0 &49.8 &9.8 &14.1 &14.3 &22.2 &3.7 &12.9 &--- &---\\
			
			Proposed &\textbf{56.1} &\textbf{96.8} &\textbf{9.2} &13.9 &\textbf{2.0} &\textbf{10.4} &\textbf{8.2} &11.0 &3.1 &\textbf{7.1} &\textbf{109.4} &\textbf{157.6}\\
			
			\noalign{\smallskip}
			\cdashline{1-13}[9pt/4pt]
			\noalign{\smallskip}

			Proposed (no-CSOE) &67.1 &105.8 &10.7 &15.9 &2.6 &10.9 &10.3 &12.5 &3.6 &7.82 &153.4 &266.3\\
			
			Proposed (no-CP) &63.9 &98.7 &10.4 &18.3 &2.9 &11.9 &8.9 &12.0 &5.3 &9.6 &112.8 &164.1\\
			
			%Proposed (no-MDCB) &81.7 &118.5 &12.7 &15.3 &3.8 &16.7 &9.5 &15.0 &4.3 &9.2 &113.4 &172.9\\
			
			Proposed (no-ARFW) &65.8 &106.9 &11.9 &13.7 &2.6 &14.7 &7.1 &12.8 &5.6 &10.6 &96.3 &165.4\\
			\hline
		\end{tabular}
	\end{center}
\end{table*}

\subsection{Crowd Counting}

As the first experiment, we conduct a comparison between our proposed approach with the state-of-the-art approaches for counting task. Quantitative results are shown in Table.~\ref{count-table}, where strong competitors (\emph{e.g.} ACSCP \cite{ACSCP}, CSR-Net \cite{CSRNet}, RAZ-Net \cite{RAZ}, etc) are included. The "Proposed" method refers to the end-to-end trainable model with CSOE+CP+MDCB+ARFW.

For the four evaluation datasets, we randomly divide their original training images into a training set (80\%) and a validation set (20\%). We perform a random grid search to tune the two hyper parameters ($m, \alpha$) of the proposed algorithm on validation set, and evaluate the algorithm on their testing set. For example, UCF-QNRF crowd dataset contains 1535 images including 1201 training images (961 training and 240 validation) and 334 testing images. After random grid search for hyper parameters, the best performance of the proposed algorithm is obtained when $m=134, \alpha=1.65$ for ShanghaiTech-A dataset; $m=125, \alpha=1.44$ for ShanghaiTech-B dataset; $m=127, \alpha=1.78$ for WorldExpo dataset; $m=153, \alpha=2.15$ for UCF-QNRF dataset.

Results on the testing sets are summarized in Table.~\ref{count-table}. On ShanghaiTech-A dataset, the proposed method gets the best performance in terms of MAE and MSE, demonstrating the effectiveness of the proposed method against outdoor scenes like the two datasets, where significant perspective variations and complex background clutter are present. On ShanghaiTech-B, the proposed method outperforms most algorithms except SA-Net (patch) and RAZ-Net as close competitors. It is necessary to mention that SA-Net (patch) performed evaluation in patch level, which is different from the standard way in the literature. When evaluated under image level, the performance of SA-Net degrades severely, this is also observed by \cite{wan2019residual}. Compared to RAZ-Net, the proposed method obtains lower RMSE. On the newest crowd dataset UCF-QNRF, the proposed method achieves remarkable performance followed by RAZ-Net. Similar to RAZ-Net, the proposed method also adopts a fusion scheme to solve the two related tasks: crowd counting and localization. While, the proposed method can use accurate crowd location information to supervise layer-wise weights optimization through the entire end-to-end training process, by introducing the compressed sensing based encoding from pixel level crowd location to robust vector representation. Thus, the proposed method further reduces the MAE and MSE level of existing counting methods (see the gap between "Proposed" and RAZ-Net).

\begin{figure}[htbp]
	\centering
	\includegraphics[width=9cm]{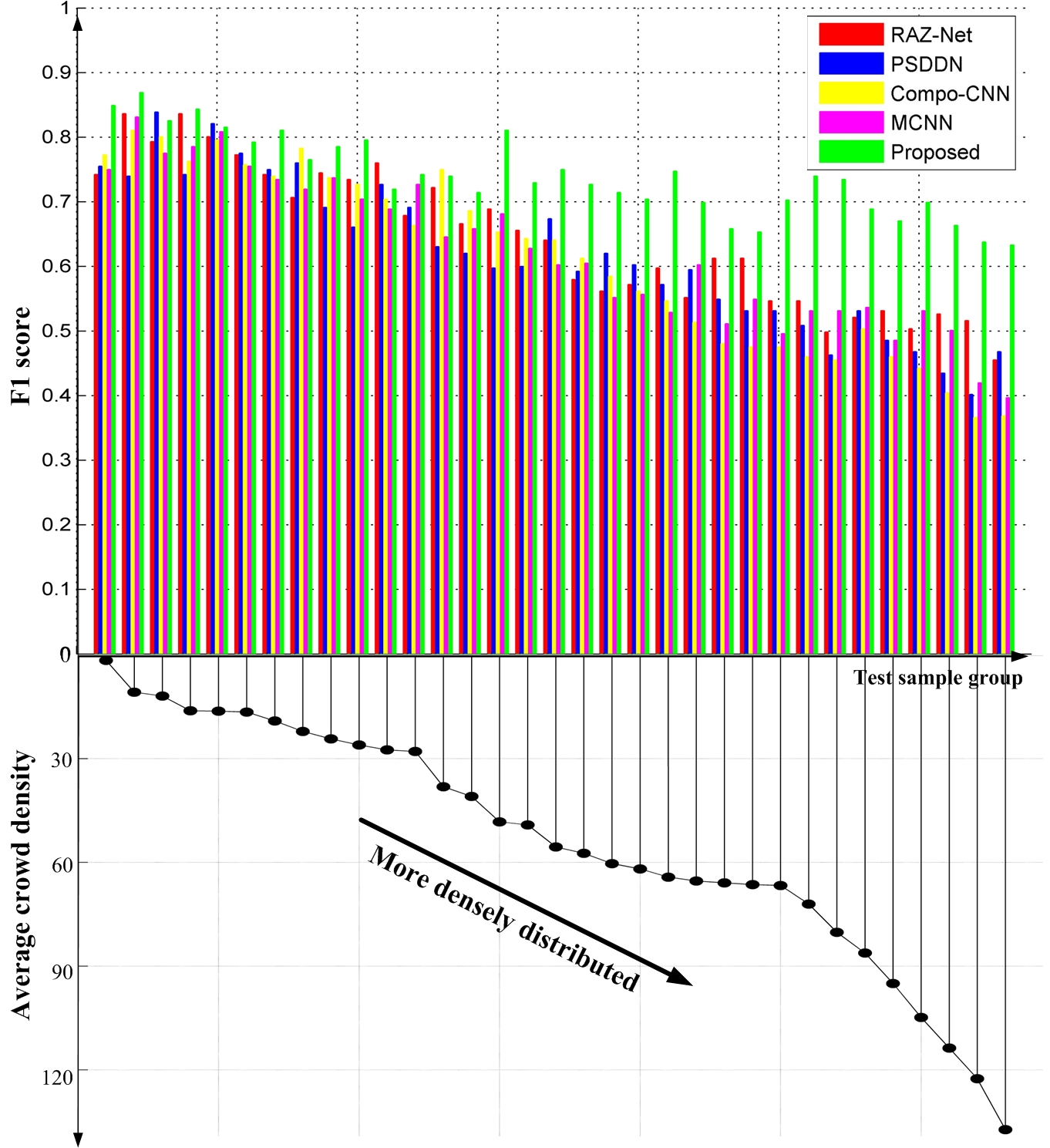}
	\caption{Detection $F_{1}$-scores with respect to average crowd density. People distribution in testing sample groups is from sparse to highly dense.}
	\label{dense-crowd}
\end{figure}

\begin{figure*}[htbp]
	\centering
	\includegraphics[width=17cm]{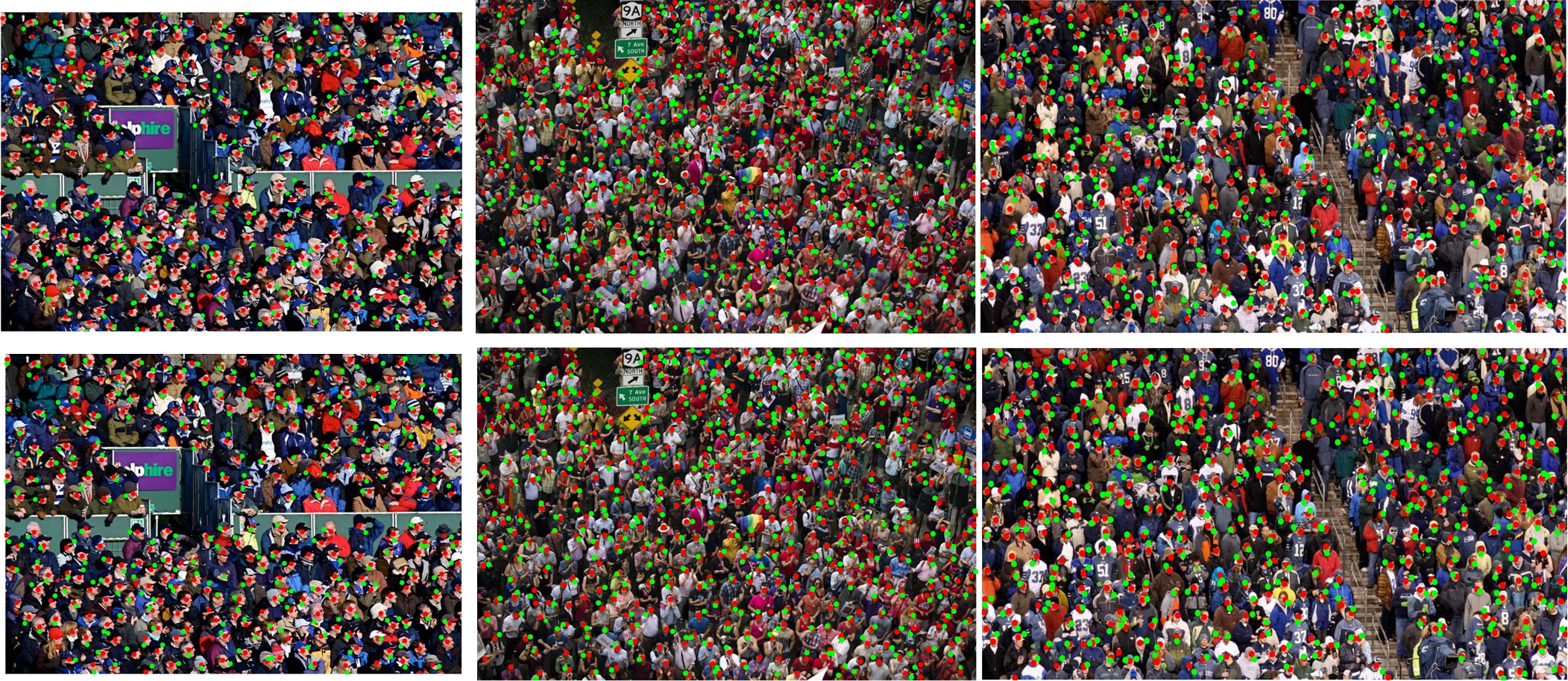}
	\caption{The red dots represent the ground truth while green dots are the locations detected by the proposed approach. \emph{Bottom} row: results are obtained by network with Compressed Sensing based Output Encoding (CSOE); \emph{top} row: results are obtained by network without CSOE. CSOE helps to boost localization performance in circumstances where targets are highly crowded but without huge scale variation.}
	\label{ablation-1}
\end{figure*}

\subsection{Crowd Localization}

\begin{figure*}[htbp]
	\centering
	\includegraphics[width=17cm]{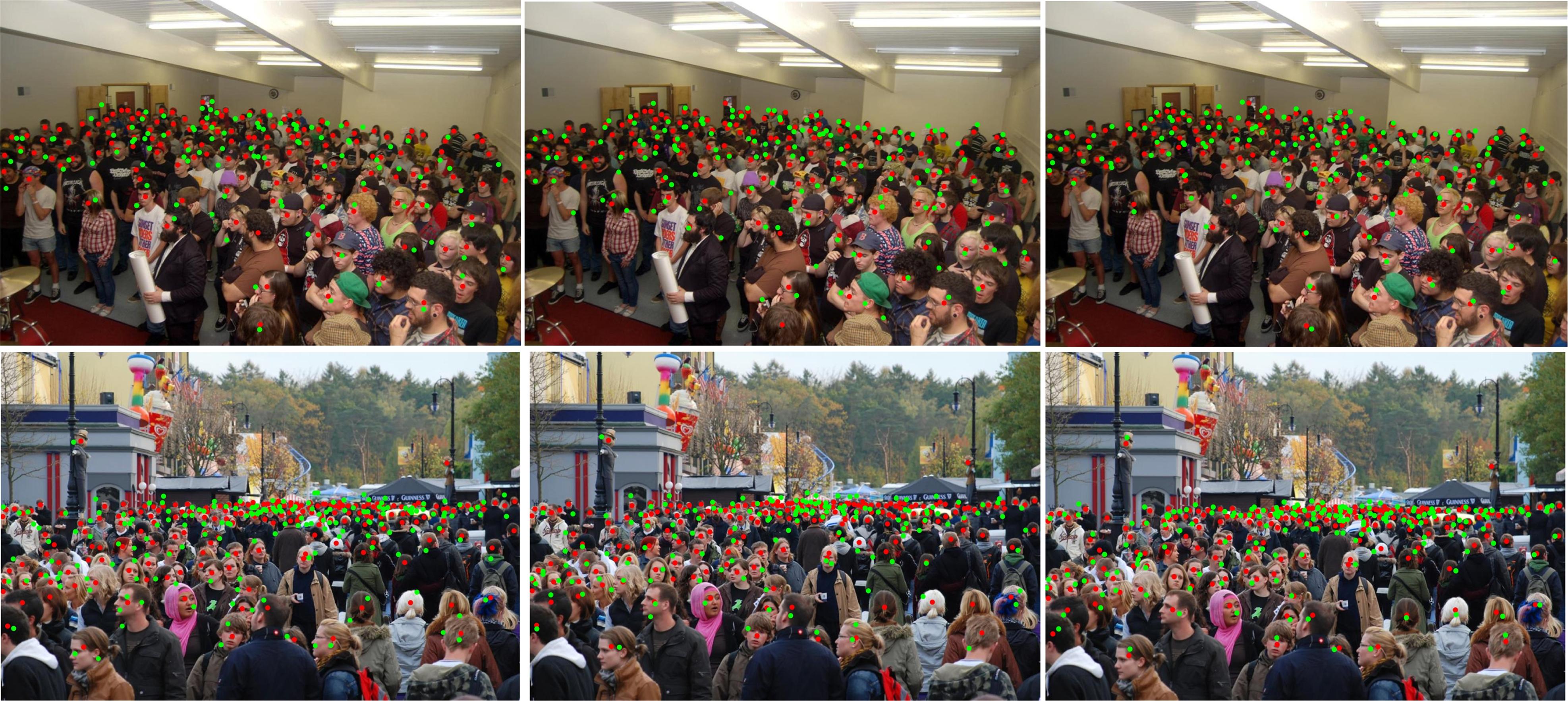}
	\caption{Red dots: the ground truth; green dots: localization results by the proposed approach. The results are obtained under the circumstance that both Compressed Sensing based Output Encoding and Center Pooling are used. \emph{Left} column: not using Multiple Dilated Convolution Branches (MDCB); \emph{middle} column: using MDCB; \emph{right} column: using both MDCB and Adaptive Receptive Field Weighting (ARFW). MDCB is good for solving huge scale variation and also avoids excessive loss of detail resolution. ARFW further deals with scale variation issue by adaptively emphasizing informative channels that have proper receptive field size.}
	\label{ablation-2}
\end{figure*}

The second experiment is a comparison between our proposed approach with state-of-the-art crowd localization methods: RAZ-Net \cite{RAZ}, Compo-CNN \cite{UCF-QNRF}, MCNN \cite{ShanghaiTech}, PSDDN \cite{BoxOut}, etc. Table.~\ref{localization-table} reports their performances in terms of Precision and Recall. On ShanghaiTech-A dataset, the proposed approach outperforms other competitors obviously in terms of both Precision and Recall. Similar to \cite{BoxOut}, we find that crowd in ShanghaiTech-A is much denser than that in ShanghaiTech-B. On ShanghaiTech-B dataset, the proposed approach is superior to RAZ-Net, Compo-CNN and MCNN. On WorldExpo dataset, the proposed approach achieves the highest Precision=0.820 and Recall=0.812. On the newest and most challenging dataset UCF-QNRF, although most methods have a slight performance decline in their Precision and Recall values, the proposed approach obtains performance improvement over RAZ-Net and Compo-CNN, which are two strong competitors. In addition, some representative localization results are shown in Fig.~\ref{ablation-1} and Fig.~\ref{ablation-2}. The red dots represent the ground truth while green dots are the locations detected by the proposed approach. It can be seen that even for very dense crowds, the proposed method still generates precise localization results.

\begin{table*}[htbp] \footnotesize
	\setlength{\abovecaptionskip}{0pt}
	\setlength{\belowcaptionskip}{0pt}
	\begin{center}
		\caption{Localization performances on four crowd benchmarks.}
		\label{localization-table}
		\begin{tabular}{*{22}{c}}
			
			\hline\noalign{\smallskip}
			
			\multirow{2}{*}{Method} &\multicolumn{2}{c}{ShanghaiTech-A} &\multicolumn{2}{c}{ShanghaiTech-B} &\multicolumn{2}{c}{WorldExpo} &\multicolumn{2}{c}{UCF-QNRF}\\
			
			&Precision &Recall &Precision &Recall &Precision &Recall &Precision &Recall\\
			\hline
			\noalign{\smallskip}

			ACSCP \cite{ACSCP} &0.792 &\textbf{0.828} &0.790 &0.601 &0.737 &0.796 &0.756 &0.597\\
			
			DecideNet \cite{DecideNet} &0.822 &0.733 &0.808 &0.788 &0.685 &\textbf{0.812} &0.593 &0.630\\
			
			Crowd-CNN \cite{Crowd-CNN} &0.819 &0.779 &0.754 &0.793 &0.738 &0.782 &0.781 &0.651\\
			
			RAZ-Net \cite{RAZ} &0.865 &0.697 &0.841 &0.758 &0.795 &0.731 &0.815 &0.711\\
			
			Compo-CNN \cite{UCF-QNRF} &0.790 &0.723 &0.781 &0.739 &0.716 &0.754 &0.717 &0.675\\
			
			MCNN \cite{ShanghaiTech} &0.765 &0.817 &0.768 &0.780 &0.724 &0.783 &0.710 &0.724\\
			
			PSDDN \cite{BoxOut} &0.760 &0.806 &0.824 &0.760 &0.809 &0.775 &0.788 &0.675\\
			
			Proposed &\textbf{0.873} &0.792 &\textbf{0.867} &\textbf{0.805} &\textbf{0.820} &\textbf{0.812} &\textbf{0.824} &\textbf{0.783}\\
			
			\noalign{\smallskip}
			\cdashline{1-9}[9pt/4pt]
			\noalign{\smallskip}
			
			Proposed (no-CSOE) &0.836 &0.745 &0.827 &0.794 &0.779 &0.786 &0.792 &0.719\\

			Proposed (no-CP) &0.851 &0.779 &0.845 &0.798 &0.801 &0.788 &0.796 &0.748\\%没有center pooling查准率降低，但是不能降太大，它的作用没那么大
		
			%Proposed (no-MDCB) &0.849 &0.734 &0.837 &0.769 &0.781 &0.730 &0.785 &0.681\\%没有Dilated Conv Branches查全率降低
			
			Proposed (no-ARFW) &0.860 &0.762 &0.852 &0.787 &0.809 &0.793 &0.804 &0.751\\
			
			\hline
		\end{tabular}
	\end{center}
\end{table*}

\textbf{Effect of High Crowd Density}

To further evaluate the existing crowd localization methods, in the third experiment, we investigate a significant issue in crowd analysis: how well the methods work in highly crowded scenes? To clarify the effect of high crowd density, we explore accuracies of five aforementioned crowd localization approaches with respect to varying crowd density. We rank test samples from WorldExpo \cite{WorldExpo} and UCF-QNRF \cite{UCF-QNRF} dataset according to the number of people present, resulting in 14080 images of size 200-by-200. We divide all the test samples into 33 groups, whose average crowd densities increase gradually from extremely sparse to extremely dense. For example: images in the first test sample group have only 1 people; images in the 15th group contain 40.8 people on average. Fig.\ref{dense-crowd} presents the F1-scores of the five crowd localization methods on the 33 test sample groups. In the first 12 groups where average crowd densities are not very high [0-30], RAZ-Net, Compo-CNN, MCNN or PSDDN achieves superior $F_{1}$-scores and outperforms "Proposed" in 4 groups. But when facing the last 21 groups whose average crowd densities are much higher, the "Proposed" obviously preserves the discrimination ability in highly crowded scenes. The relative $F_{1}$-scores gains over the 4 methods increase with a higher average crowd density. More specifically, it is in the range of average crowd density = [60-135] (\emph{i.e.} 20th to 33th group) that Compo-CNN, MCNN and PSDDN show rapid and obvious performance declines suffering from the increasing average crowd density. In the most crowded group, the $F_{1}$-scores of RAZ-Net, MCNN and PSDDN have drop to the range of 0.40-0.48; in comparison "Proposed" maintains a $F_{1}$-score=0.637. The trend is clear. As the crowd density increases the accuracy gap between "Proposed" and other methods increases, supporting our claim that regression in encoding signal space is better than detecting pixel coordinates of small objects in pixel space for crowd analysis in highly congested scenes.

\subsection{Ablation study}
We carry out ablation experiments to better understand the effect of four major components of the proposed method.

\textbf{Compressed Sensing based Output Encoding (CSOE)}

To investigate the role of CSOE, we design a comparison neural network which directly predicts the heatmap of the head centroid position without the use of CSOE module and reconstruction module. Figure.~\ref{ablation-1} shows localization results obtained by network with CSOE (bottom) and without CSOE (top). The "Proposed (no-CSOE)" row in Table.~\ref{count-table} and Table.~\ref{localization-table} gives the counting results and localization results of not using CSOE on four datasets. Specifically speaking, CSOE helps to boost counting and localization performance in highly crowded circumstance, especially for cases where the density of targets is high but without huge scale variation. It can be also observed that the big "precision" gap of "Proposed" over "Proposed (no-CSOE)".

\textbf{Center Pooling (CP)}

The "Proposed (no-CP)" row in Table.~\ref{count-table} and Table.~\ref{localization-table} gives the counting results and localization results of not using center pooling on four datasets. Center pooling slightly improve localization accuracy, especially in terms of Precision. Center pooling is good for mining the recognizable features within objects.

\textbf{Multiple Dilated Convolution Branches (MDCB)}

%The "Proposed (no-MDCB)" row in Table.~\ref{count-table} and Table.~\ref{localization-table} gives the counting results and localization results of not using MDCB on four datasets.
MDCB helps to improve localization accuracy when head size changes drastically, since the Multiple Branches offer a set of different receptive field sizes. In addition, unlike conventional convolution+pooling operation, dilated convolution will not bring excessive loss of detail resolution.

\textbf{Adaptive Receptive Field Weighting (ARFW)}

The "Proposed (no-ARFW)" row in Table.~\ref{count-table} and Table.~\ref{localization-table} gives the counting results and localization results of not using ARFW on four datasets. To further handle the issue of huge scale variation in an image, we propose Adaptive Receptive Field Weighting (ARFW), which enables our model to use global information to adaptively emphasize informative channels that have proper receptive field size. Figure.~\ref{ablation-2} depicts localization results obtained by three models with different configurations: (1) left column: CSOE+CP only; (2) middle column:  CSOE+CP+MDCB; (3) right column: CSOE+CP+MDCB+ARFW.

As we know, the receptive field size of baseline model (CSOE+CP) is fixed. It is often observed that such fixed receptive field size is hard to balance between small targets and large targets. Take 2nd row as an example, we speculate the receptive field size of baseline is relatively small. Consequently the localization performance of distant small targets is okay, but multiple localizations occur on one large target that is close to camera. However, with the introduction of MDCB and ARFW, the complete model (CSOE+CP+MDCB+ARFW) has a set of different receptive field sizes and is capable of adaptively emphasizing the proper receptive field size.

\textbf{Components Configuration}

Table.\ref{configuration-experiment} presents $F_1$-scores of the proposed method using different configurations on the four crowd benchmarks. Here CSOE, MDCB, CP and ARFW are used separately or jointly to fully explore their influences to localization performances. Not surprisingly, the complete model CSOE+MDCB+CP+ARFW achieves the best results. We can observe that CSOE+MDCB+CP+ARFW performs much better than CSOE+MDCB+CP and MDCB+CP+ARFW. This demonstrates that both CSOE and ARFW are crucial and bring huge contributions to overall performance. While the small gap between CSOE+MDCB+CP+ARFW and CSOE+MDCB+ARFW indicates that CP can bring slight benefits to performance, similar to the finding in \cite{CenterNet}. It is also intesting to see that MDCB+ARFW acquire the highest $F_1$-scores among all four "two-tricks" configurations: CSOE+MDCB, CSOE+CP, MDCB+CP and MDCB+ARFW. This phenomenon provides strong evidence to the importance of tackling target size variation issue in crowd analysis task. While CSOE is the best among all "single-trick" configurations. Please note that ARFW must work with MDCB, so there is no configuration where only ARFW is used. The better performance of CSOE further supports our previous claim that regression in encoding signal space is better than detecting pixel coordinates of small objects in pixel space for crowd analysis in highly congested scenes.

\begin{table}[h] \footnotesize
	\setlength{\abovecaptionskip}{0pt}
	\setlength{\belowcaptionskip}{0pt}
	\begin{center}
		\caption{Localization performances ($F_1$-score) of the proposed method using different configurations on the four crowd benchmarks. Please note that ARFW must work with MDCB, so there is no configuration where only ARFW is used.}
		\label{configuration-experiment}
		\begin{tabular}{*{22}{c}}
			\hline\noalign{\smallskip}
			CSOE & MDCB & CP & ARFW & Sh-A & Sh-B & Wor & UCF\\
			\noalign{\smallskip}
			\hline
			\noalign{\smallskip}
			
			\checkmark &  &  & &0.695 &0.689 &0.677 &0.665\\
			
			&\checkmark & & &0.684 &0.670 &0.665 &0.652\\
			
			& &\checkmark & & 0.637 &0.610 &0.608 &0.571\\
			
			\checkmark & \checkmark&  & &0.761 &0.752 &0.697 &0.762\\
			
			\checkmark & & \checkmark& &0.748 &0.732 &0.720 &0.711\\
			
			& \checkmark & \checkmark& &0.723 &0.738 &0.690 &0.709\\
			
			& \checkmark&  &\checkmark &0.787 &0.802 &0.755 &0.729\\
			
			\checkmark & \checkmark& \checkmark& &0.808 &0.818 &0.801 &0.777\\
			
			& \checkmark & \checkmark&\checkmark &0.788 &0.810 &0.783 &0.754\\
			
			\checkmark& \checkmark & &\checkmark &0.823 &0.821 &0.794 &0.791\\
			
			\checkmark& \checkmark &\checkmark &\checkmark&0.831 &0.834 &0.816 &0.803\\
			
			\hline
		\end{tabular}
	\end{center}
\end{table}

\section{Conclusion}
In this paper we developed several core modules CSOE, MDCB, ARFW and sparse reconstruction layers, and also integrate them into an end-to-end trainable network. A wide range of experiments show the effectiveness of the proposed method, which presents state-of-the-art performance across multiple datasets, especially achieves excellent results in scenes with high crowd density. Experiments support our insights that it is crucial to tackle target size variation issue in crowd analysis task, and casting crowd localization as regression in encoding signal space is quite effective in crowd scenes. We hope these insights may prove useful for other crowd analysis tasks.

\ifCLASSOPTIONcaptionsoff
  \newpage
\fi



\bibliographystyle{IEEEtran}
\bibliography{refs}

\end{document}